\documentclass[letterpaper]{article}

\usepackage{natbib,alifeconf}  
\usepackage{url,graphicx,subcaption, hyperref,cleveref,amsmath,mathtools,amssymb}
\usepackage{booktabs}

\usepackage{lipsum}
\usepackage{xcolor}
\usepackage[normalem]{ulem}
\newcommand\blfootnote[1]{%
  \begingroup
  \renewcommand\thefootnote{}\footnote{#1}%
  \addtocounter{footnote}{-1}%
  \endgroup
}

\title{Implementation of reinforcement learning in chemical reaction networks: application to phototaxis as curiosity-driven exploration}

\author{
    Ruyi Tang$^{1}$,
    Grégoire Sergeant-Perthuis$^{1, *}$, \and
    David Colliaux$^{2, *}$ \\
    \mbox{}\\
    $^1$CQSB, Sorbonne University, France\\
    $^2$Sony CSL, France\\
   evatang2002@sina.com
} 

\begin{document}

\maketitle

\begin{abstract}
    
    Living systems navigate environments using noisy and incomplete sensory signals. In unicellular algae, phototaxis is often modeled as a mechanistic run–tumble process driven by stimulus–response rules. However, such descriptions overlook how organisms actively sample their environment to reduce sensory ambiguity. From a minimal cognition perspective, we reframe this navigation as a subjective, information-driven sensorimotor process. 

    To this end, we propose a framework linking a Partially Observable Markov Decision Process (POMDP) with biochemical reaction dynamics. Environmental variables are hidden, while the cell updates a minimal internal state from each observation through a memoryless Bayesian step. These internal dynamics balance orienting toward light with exploratory reorientation and can be implemented through Chemical-Reaction-Network Ordinary Differential Equations (CRN–ODEs). Our model includes a biophysical observation process for photoreception and a chemically computable polynomial bound on information gain. Using Inverse Reinforcement Learning (IRL) on 30 experimentally recorded Chlamydomonas trajectories, we infer the behavioral objective consistent with observed phototactic motion and benchmark the resulting dynamics with standard Stochastic Simulation Algorithm (SSA) baselines. Our model reproduces the empirical alignment-to-light distribution, comparable to objective SSA baselines on this dataset.  Within this framework, run–tumble alternation emerges as an information-acquisition strategy: tumbling reorients the cell to sample new sensory configurations and resolve sensor ambiguity, demonstrating how intracellular biochemical networks can support adaptive information-seeking behavior in cellular navigation.

\end{abstract}


Data/Code available at: \url{https://github.com/giveyourselfaTRY/phototaxis-pomdp-crn}

\blfootnote{\textcopyright\  2026 Ruyi Tang,
    Grégoire Sergeant-Perthuis,
    David Colliaux. Published under a Creative Commons Attribution 4.0 International (CC BY 4.0) license.
    
   $^*$ Authors contributed equally.}

\section{Introduction}

Across biological scales—from insects tracking turbulent plumes~\citep{heinonen2025optimal} to bacteria climbing microscopic gradients~\citep{auconi2022gradient}—organisms navigate complex environments using noisy and incomplete sensory cues. In minimal organisms, such navigation emerges without neural systems, raising a central question in artificial life and minimal cognition: how can adaptive behavior arise from simple biochemical dynamics interacting with uncertain environments? In this paper, we provide a plausible explanation of the behavioural role of tumbling in phototaxis; namely, that it is a way, in the context of ambiguous sensory input, to resolve the ambiguity and ensure optimal reward.

Recent work suggests that cellular signaling networks can function as distributed information-processing systems~\citep{COLLIAUX2017385}. At the molecular level, biochemical reaction networks are capable of implementing probabilistic computations, including Bayesian-like inference, through populations of interacting molecules and diffusible messengers~\citep{doi:10.1098/rsif.2022.0877}. These results point toward a possible bridge between statistical models of inference and the physical dynamics of intracellular chemistry. At the behavioral scale, navigation can also be interpreted as an information-driven sensorimotor process, where organisms actively sample their environment to reduce uncertainty---a principle closely related to active sensing~\citep{gottlieb-2018}. To mathematically formalize this mechanism, Reinforcement Learning (RL) offers a powerful paradigm. Specifically, the concept of \textit{curiosity} in RL is formally defined as an intrinsic reward signal proportional to the expected information gain (or reduction in uncertainty) about the environment's latent states~\citep{8014804}. However, how such curiosity-driven, information-seeking dynamics could arise from biochemically realizable mechanisms remains comparatively underexplored.

Whereas prior work treats these ingredients separately---CRN-based stimulus--response and operant-conditioning-like adaptation~\cite{10.1371/journal.pcbi.1010676}, chemical learning of generative models~\cite{10.1098/rsif.2024.0373}, and active-inference accounts of microbial navigation~\cite{tschantz2020learning}---our contribution is to integrate them in the algal phototaxis setting as an adaptive curiosity-driven decision-making process under partial observability. We formulate the navigation problem as a Partially Observable Markov Decision Process (POMDP) and demonstrate that its internal inference and decision dynamics can be implemented using modular Chemical-Reaction-Network Ordinary Differential equations (CRN–ODEs). This connects RL prescribed policies and dynamics with potential biochemical processes.

\subsection{Contributions}

The novelty of this work lies in its phototaxis-specific integration rather than in any single ingredient alone. Prior work has studied CRN-based decision-making, autonomous chemical learning, and epistemic action in active inference. Here, we bring these ideas into a concrete biological navigation problem by linking photoreceptor geometry, subjective POMDP inference, data-driven IRL, and CRN--ODE implementation in a single pipeline.

Specifically, we make three contributions. First, we formulate \textit{Chlamydomonas} phototaxis as a subjective POMDP driven by coarse, biophysically motivated left--right observations, thereby making the front--back ambiguity induced by eyespot geometry explicit. Second, we use 30 experimentally recorded trajectories to learn a data-driven phototactic policy and recover a reward structure through inverse reinforcement learning, and we compare the resulting dynamics with stochastic mechanistic baselines using alignment-distribution diagnostics. Third, we construct a CRN--ODE realization of the one-step belief update and action-evaluation modules, including a polynomial surrogate for information gain that is compatible with mass-action kinetics. These components provide an end-to-end bridge from sensory geometry and subjective inference to biochemical implementation and empirical trajectory statistics, giving a quantitative explanation of how tumbling supports phototaxis rather than acting as stochastic motor noise.


Together, these results support the view that intracellular biochemical networks could implement components of adaptive information-seeking behavior, linking cellular chemistry, reinforcement-learning-inspired information processing, and minimal cognition in microbial navigation.

\section{From objective to subjective probabilistic model for phototaxis}

\subsection{An objective model for mechanistic run-and-tumble motion.}
To establish a behavioral baseline, we first formulate an idealized ``objective'' model of phototaxis, implemented via the standard Stochastic Simulation Algorithm (SSA)~\citep{doi:10.1021/j100540a008}. Modeling run-and-tumble navigation as a stochastic velocity-jump process governed by environment-dependent transition rates is widely validated in theoretical biology~\citep{doi:10.1137/S0036139903433232, article, polin2009chlamydomonas}. This mechanistic framework adopts an omniscient observer's perspective, assuming the dynamics are driven strictly by exact, global physical quantities.

For the objective model, we consider the organism to change orientation at random times with a rate increasing with the misalignment of its motion relative to the light source. We suppose the cell swims in a straight line in-between those tumbling events. Specifically, at any time $t$, the agent's spatial state is defined by its position $x_t$ and heading vector $\hat{u}_t$. The objective misalignment angle $\theta$ is derived from the true alignment cosine between the heading and the absolute light source $\mu$: $\cos\theta = \hat{u}_t \cdot (\mu - x_t) / \|\mu - x_t\|$.

Following the classical instantaneous reorientation framework, the time interval $\Delta t$ between two tumbling events is drawn from an exponential distribution $\Delta t \sim \text{Exp}(\lambda(\theta))$. To capture the mechanistic intuition that tumbling probability increases as the agent deviates from the target, we propose the following penalty function for the tumbling rate:
\begin{equation}
    \lambda(\theta) = \lambda_{\min} + (\lambda_{\max} - \lambda_{\min}) \left( \frac{1 - \cos\theta}{2} \right)^\gamma
    \label{eq:lambda_rt}
\end{equation}
During the drawn interval $\Delta t$, the agent executes a \textit{run advance}. Its position is explicitly updated via the kinematic equation $x_{t+\Delta t} = x_t + c_{\text{run}} \hat{u}_t \Delta t$, while its heading $\hat{u}_t$ is subject to marginal rotational noise $\mathcal{N}(0, \sigma^2)$. Once the tumbling event is triggered, the reorientation is assumed to be instantaneous: the agent immediately samples a new heading uniformly from all possible directions, $\hat{u}_{\text{new}} \sim \text{Unif}(\mathbb{S}^1)$, without any translational displacement. 





While this objective model reproduces macroscopic gradient-ascent trajectories, it assumes the cell has a built-in global navigator that computes the exact misalignment $\theta$ to evaluate Equation \ref{eq:lambda_rt}. In reality, algae decide using only localized, noisy photon fluxes on the eyespot. We therefore shift from objective mechanics to subjective inference.

\subsection{A subjective model for the curiosity-driven directed run-and-tumble motion.}

\subsubsection{Partial observability: biophysical model}
To model the subjective perspective of the cell, we first define its sensory limitations. Unlike the objective model which assumes access to global alignment, real algae rely on localized receptors. Although the real alga (\textit{Chlamydomonas reinhardtii}) possesses a single eyespot and swims while spinning in 3D, we adopt a 2D projection model to capture its sensory dynamics. The incident light flux is decomposed into left and right channels relative to the current heading; these ``two eyes'' are a modeling surrogate for directionality and body occlusion in 2D, not an anatomical claim.

\begin{figure}[htbp]
    \centering
    \includegraphics[width=0.95\linewidth]{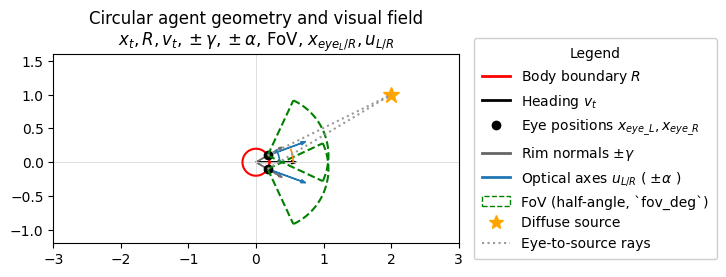}
    \caption{Illustration of the biophysical agent in 2D. The directional shielding and limited field-of-view of the sensors create blind spots and geometric ambiguity regarding the true light source position.}
    \label{fig:biophy}
\end{figure}

As illustrated in Figure \ref{fig:biophy}, we model the agent as a circular body of radius $R>0$, located at $x_t \in \mathbb{R}^2$ with a unit heading vector $v_t \in \mathbb{S}^1$. Two virtual sensors are mounted on the body rim at symmetric offset angles $\pm \gamma$ relative to the heading. Furthermore, the optical axes of these sensors are directed at angles $\pm \alpha$ relative to $v_t$, and each sensor can only perceive incoming light within a restricted half field-of-view ($FoV$).

This physical embodiment fundamentally acts as a sensory bottleneck. As we will formalize in the subsequent POMDP framework, this restricted sensor configuration induces inherent \textit{geometric ambiguity}. Specifically, due to the Field-of-View (FoV) limits of each sensor, when the light source lies directly behind the agent or directly in front, the alga receives essentially the same low sensory input on its lateral photoreceptors, resulting in minimal activation that obscures the true light direction.

\subsubsection{POMDP}
Given the sensory bottleneck and the resulting partial observability, we cast the phototaxis navigation as a subjective inference process by formulating a Partially Observable Markov Decision Process (POMDP).

\paragraph{States and actions:} The hidden state $h \in \{0,\dots,n-1\}$ (with $n=5$ in our experiments) discretizes the relative angle between the agent's heading and the light source. Each $h$ is associated with an angle \(\theta(h)=\frac{2\pi h}{n}\). The action space $\mathcal{A}=\{0,1\}$ dictates movement: $a=0$ (run) maintains the current heading with forward displacement, while $a=1$ (tumble) pauses forward motion to randomly sample a new heading from a uniform distribution.

\paragraph{Observations:} 
At each time step $t$, the agent receives a discrete observation $o_t = (o_L, o_R) \in \{0, \dots, m-1\}^2$. To explain how this observation is calculated from the spatial geometry, let $\theta$ be the relative angle to the light source. The incident light intensity $I_c$ on each sensor $c \in \{L, R\}$ is proportional to its optical axis alignment $\max(0, \cos(\theta \mp \alpha))$ and is explicitly gated by the \textit{FoV}. To model the non-linear photoreceptor physics, this raw signal is passed through a logistic saturation function and quantized into $m$ discrete levels (set to $m=2$, yielding 4 distinct observation pairs):
\begin{equation}
o_c = \text{bin}\left( \frac{1}{1 + \exp(-\beta_s (I_c - I_{50}))} \right)
\label{eq:observation}
\end{equation}
where $\beta_s$ controls the saturation steepness and $I_{50}$ is the half-response threshold.


\paragraph{Transition and emission:}



The transition matrix $T_a(h'|h)$ encodes the rotational dynamics. A \textit{run} preserves the relative angle with marginal diffusion leakage $\epsilon$, yielding a diagonally dominant matrix: $T_{\text{run}}(h'|h) = \frac{\delta_{h',h} + \epsilon}{1 + \epsilon n}$. A \textit{tumble} completely randomizes the heading, yielding a uniform transition: $T_{\text{tumble}}(h'|h) = 1/n$.

The emission kernel $Z(o|h) = P(o|h)$ mathematically bridges the abstract hidden states and the physical observations. To construct this matrix, each hidden state $h$ is assigned a prototype angle $\theta_h = \frac{2\pi h}{n}$. We pass $\theta_h$ through the biophysical observation model (Eq. \ref{eq:observation}) to determine its deterministic target observation $o^*$. We then assign a high probability spike to $o^*$ and a uniform noise floor $\epsilon_{\text{obs}}$ to all other bins, followed by column normalization. This yields a robust, physically-grounded likelihood matrix.

\paragraph{Memoryless belief update:}
The agent maintains a subjective belief $b_t$, a probability distribution over the hidden states $h$. We formulate a memoryless Bayesian update so that it can be implemented by 
executable biochemical dynamics. At each step $t$, the posterior belief $b^+_t$ is computed via a single-step likelihood reweighting over a fixed uniform prior $b_0$:
\begin{equation}
b^+_t(h) = \frac{Z(o_t|h) b_0(h)}{\sum_{j=0}^{n-1} Z(o_t|j) b_0(j)}
\end{equation}

\subsubsection{Curiosity-driven one-step-look-ahead policy}
To select action based on subjective memoryless belief, we adopt a  one-step look-ahead policy, scoring each candidate action by balancing task-oriented exploitation and information-gathering exploration. After contemplating an action $a$, the anticipated next-state belief is $b'_a = T_a b^+_t$. The total value of an action is defined as:
\begin{equation}
V(a) = V^{(1)}(a) + \lambda V^{(2)}(a)
\end{equation}
where $\lambda \ge 0$ is a tunable hyperparameter balancing the exploration-exploitation trade-off. The exploitation term $V^{(1)}$ represents the expected immediate task reward:
\begin{equation}
V^{(1)}(a) = \mathbb{E}_{h' \sim b'_a}[r] = r^\top (b'_a)
\end{equation}
where the reward vector $r \in \mathbb{R}^n$ assigns higher values to hidden states that are closely aligned with the light source, which is why we take the global alignment into account during the next-stage learning. This exploitation term plays a role analogous to instrumental value in active inference, but it is not identical to it: instrumental value is an expected log-preference over future observations, whereas here preferences are encoded directly as rewards over latent alignment states.
The exploration term $V^{(2)}$ measures the expected information gain about the hidden state $H$ from the future observation $O$ predicted under action $a$. We quantify this using Mutual Information (MI), which is mathematically equivalent to the one-step state-epistemic-value component of expected free energy~\citep{Friston02102015}, here in the fixed-parameter form of ~\cite{tschantz2020learning}:
\begin{equation}
V^{(2)}(a) = I(H; O \mid b'_a) = \sum_{o, h} b'_a(h) Z(o|h) \log \frac{Z(o|h)}{p_O(o)}
\end{equation}
where $p_O(o) = \sum_h Z(o|h) b'_a(h)$. Model parameters are fixed during navigation, so we do not include the parameter-epistemic-value term used for online generative-model learning in active inference. Finally, actions are selected either via a Boltzmann policy, $\pi(a) \propto \exp(V(a)/\tau)$, where the temperature $\tau$ modulates the stochasticity of action sampling; or deterministically via $\arg \max_a V(a)$. Thus, hyperparameters $\lambda$ and $\tau$ play distinct roles: $\lambda$ changes the action values by weighting the information-gain term, whereas $\tau$ only changes how sharply the policy samples from the resulting values.

By formulating this subjective policy, we hypothesize that the curiosity term $V^{(2)}$ might drive the agent to actively resolve the geometric ambiguity highlighted in the previous section. Whether this theoretically translates to functional tumbling behavior will be empirically examined in our simulation results.

\section{Implementation in chemical reaction networks}
\subsection{Polynomial upper bound of curiosity.}

Recent theoretical frameworks demonstrate that biochemical cascades can function as ``probabilistic computers,'' where mass-action kinetics at steady state can compute any Rational Function with Non-negative Coefficients (RFNC)~\citep{COLLIAUX2017385}.

While the exploitation term $V^{(1)}$ is intrinsically linear and thus directly computable by Chemical Reaction Networks (CRNs), the exploration term $V^{(2)}$ requires computing logarithms. Since the logarithm is not an RFNC, it cannot be directly evaluated by mass-action polynomials.

To obtain a chemically implementable surrogate, we upper-bound the mutual-information term by applying the algebraic inequality $\log x \le x - 1$ to the logarithmic factor and then collecting the action-independent constant terms. This yields the following RFNC-compatible polynomial form:
\begin{equation}
V^{(2)}(a) \le \sum_{o, h} b'_a(h) Z(o|h)^2 + m^2 - 2
\end{equation}
Since $m^2 - 2$ is constant across all actions, it does not affect the action-selection policy. Thus, we substitute $V^{(2)}(a)$ with a fully chemically computable polynomial curiosity score:
\begin{equation}
\tilde{V}^{(2)}(a) = \sum_{h} b'_a(h) \sum_{o} Z(o|h)^2
\label{eq:crn_curiosity}
\end{equation}
This pure RFNC formulation allows us to seamlessly construct the corresponding CRN-ODE modules.

\subsection{CRN implementation.}

Systematic CRN compilers have been developed to program diverse input functions. Classic strands include deterministic computation of semilinear functions (\cite{chen2013deterministicfunctioncomputationchemical}) and the synthesis of CRNs for finite-support discrete probability distributions (\cite{cardelli2018programmingdiscretedistributionschemical}). More recently, \cite{doi:10.1098/rsif.2022.0877} realized dynamic inference by translating Hidden Markov Models (HMMs) into CRNs that simulate the Baum-Welch algorithm via mass-action kinetics.

Finding a middle ground between static CRN compilers and complex dynamic HMM networks, our CRN architecture integrates the monotone-accumulation principles of static computation with a forward-style inference scaffold. We construct parallel sub-networks for each action $a \in \{0(\text{run}), 1(\text{tumble})\}$, which share the same structural motifs but differ in reaction rates parameterized by $T_a$. The network operates through mass-action Ordinary Differential Equations (ODEs) across three modular blocks:

\subsubsection{1. Memoryless one-step belief update:} 
Let $A^{(1)}_h$ denote the species storing the prior belief $b_0(h)$, and $E_v$ be a boolean catalyst activated by receiving observation index $v$. The unnormalized likelihood $U_h$ is accumulated catalytically:
\begin{equation}
A^{(1)}_h + E_v \xrightarrow{k_{v,h}} A^{(1)}_h + U_h + E_v \quad (k_{v,h} = Z[v, h])
\end{equation}
To execute Bayesian normalization without complex division circuits, we introduce a shared resource pool $M$ (initialized to $1.0$). The posterior belief species $A^{(2)}_h$ is produced by consuming $M$ proportionally to $U_h$:
\begin{equation}
M + U_h \xrightarrow{1} A^{(2)}_h + U_h.
\end{equation}

This reaction must be considered jointly with the depletion of the shared pool,
\begin{equation}
    \frac{d[M]}{dt}=-k[M]\sum_j [U_j],
\qquad
\frac{d[A_h^{(2)}]}{dt}=k[M][U_h].
\end{equation}
Since the $U_h$ are fixed during one observation step, all species $A_h^{(2)}$ compete for the same normalization resource $M$. Hence
at steady state as $t \to \infty$ and $M \to 0$, the concentration $[A^{(2)}_h]$ cleanly converges to the normalized posterior $b^+_t(h)$. 
\begin{equation}
    [A_h^{(2)}]_{\infty}
=
\frac{[U_h]}{\sum_j [U_j]}
=
\frac{Z[o_t,h]\,b_0(h)}{\sum_j Z[o_t,j]\,b_0(j)}
=
b_t^{+}(h).
\end{equation}
For a POMDP with $n$ hidden states and $m^2$ possible observation pairs, this modular design scales efficiently, requiring only $\mathcal{O}(n + m^2)$ chemical species and $\mathcal{O}(n \cdot m^2)$ reactions. 



\subsubsection{2. Exploitation (immediate expected reward):} 
To compute $V^{(1)}(a) = r^\top T_a b^+_t$, we employ a ``leaky reservoir'' motif. A reward species $R$ is catalytically produced by the posterior $A^{(2)}_h$ and simultaneously degrades:
\begin{equation}
A^{(2)}_h \xrightarrow{w_a(h)} A^{(2)}_h + R, \quad R \xrightarrow{1} \emptyset
\end{equation}
where the production rate is $w_a(h) = (r^\top T_a)_h$. By solving the mass-action ODE $\dot{R} = 0$, the steady-state concentration $[R]_\infty$ exactly yields the exploitation score $V^{(1)}(a)$.

\subsubsection{3. Exploration (curiosity polynomial upper bound):} 
Similarly, to compute the polynomial bound $\tilde{V}^{(2)}(a)$ (Eq. \ref{eq:crn_curiosity}), we introduce a curiosity species $C_h$ for each state, using the transitioned likelihood squared as the production rate:
\begin{equation}
A^{(2)}_h \xrightarrow{c_h} A^{(2)}_h + C_h, \quad C_h \xrightarrow{1} \emptyset
\end{equation}
where $c_h = (T_a)_{h,:} \sum_v Z[v, \cdot]^2$. 

Ultimately, the total CRN-ODE action value is read out at steady state as $V_{\text{CRN}}(a) = [R]_\infty + \lambda \sum_h[C_h]_\infty$. The motor behaviour emerges from the competition between the two action species: an action is selected with probability proportional to their relative steady-state concentrations $V_{\mathrm{CRN}}(\text{run})$ and $V_{\mathrm{CRN}}(\text{tumble})$, a chemical analogue of the Boltzmann policy.

\section{Simulations \& comparison to data}
To rigorously evaluate our framework, we divide our experiments into two stages. First, we provide simulations for phototactic behavior under a single light source to  verify the mathematical and biochemical equivalence of our CRN design. Second, we transition to data-driven learning, extracting latent policies from real biological trajectories and benchmarking them against objective mechanistic models.

\subsection{Numerical \& CRN POMDP simulations.}
We first validate the computational integrity of our subjective POMDP and its CRN-ODE implementation in an idealized single-source environment. The agent is placed in a simulated 2D field with a distant light source and operates under a fixed, hand-specified reward vector designed to favor light alignment. 

Our goal here is to verify that the mass-action CRN-ODE faithfully reproduces the numerical policy. We simulated trajectories using both exact numerical computations and numerical integration of the CRN-ODEs (running until steady-state convergence at each decision step).

\begin{table}[htbp]
\centering

\resizebox{\columnwidth}{!}{%
\begin{tabular}{lll}
\toprule
\textbf{Simulation} & \textbf{Curiosity Weight ($\lambda$)} & \textbf{Tumble Ratio} \\ \midrule
numerical, exact MI & 0.3  & $9/4000 \approx 2.3\times10^{-3}$     \\
    numerical, exact MI  & 6.83 (coarsely-tuned) &  $29/4000 \approx 7.3\times10^{-3}$   \\
    numerical, MI upper bound     & 6.83       & $7/2000 = 3.5\times10^{-3}$   \\
    CRN--ODE      & $6.83$  & $4/2000 = 2.0\times10^{-3}$       \\
    numerical, MI upper bound     & 2(grid searched)       & $2/2000 = 1.0\times10^{-3}$   \\
    numerical, soft policy      & $0$  & $7/2000 = 3.5\times10^{-3}$       \\
    CRN--ODE, soft policy      & $0$  & $10/2000 = 5.0\times10^{-3}$       \\
\bottomrule
\end{tabular}%
}
\caption{Tumble ratio for different policy simulations and curiosity weights with single-light}
\label{table:single-sim}
\end{table}


As summarized in Table ~\ref{table:single-sim}, the tumble ratios (calculated as the fraction of tumble steps over the entire trajectory) are consistently maintained at a low magnitude ($\sim 10^{-3}$) across all models. This indicates that all configurations successfully guide the agent toward the light source with high locomotive efficiency. Beyond the aggregate tumble ratio, we verified that the surrogate preserves the decision
itself: across $6{,}506$ observation-induced, on-trajectory, and random beliefs,
$\arg\max_a \tilde V(a)$ matched $\arg\max_a V(a)$ in $99.1\text{--}99.5\%$ of cases
($100\%$ at $\lambda=0$ and for all on-trajectory beliefs), with disagreements confined to
near-tie random beliefs. The polynomial form of mutual information term is therefore a chemically computable
approximation that, while not order-preserving in general, preserves the policy over the
visited regime.

Furthermore, we explicitly distinguish between the discrete \textit{numerical} POMDP (which computes exact matrix multiplications instantaneously) and the \textit{CRN-ODE} implementation (which relies on continuous-time integration of mass-action kinetics until steady-state). The CRN-ODE yields tumbling statistics ($2.0 \times 10^{-3}$) that closely track its numerical counterpart ($3.5 \times 10^{-3}$). The slight numerical divergence arises naturally from the finite integration time in ODE solvers compared to instantaneous discrete math. Nonetheless, this confirms that the complex Bayesian belief updating and curiosity evaluation can be robustly executed by physical biochemical reaction networks.

\paragraph{Key insight: tumbling as active exploration under geometric ambiguity.} Recall the geometric ambiguity from the sensory bottleneck: the eyespot sits laterally and slightly behind the cell, so light sources at certain off-axis angles in front of or behind the cell yield near-indistinguishable bilateral readings. In these regimes the posterior over the hidden angle flattens and spatial uncertainty rises, so the curiosity term assigns greater value to information-gaining actions. Tumbling then emerges as an active exploration step: randomizing the heading changes the illumination perspective and gathers the evidence needed to resolve the front--back ambiguity. Run--tumble alternation is thus not motor noise but an active-sensing strategy for partial observability.

\subsection{Data-driven policy learning.}
The above simulations establish the computational feasibility of our CRN-POMDP framework for phototaxis scenarios. However, establishing true biological validity requires learning the policy from real organism trajectories. Our goal in this section is to train the policy on experimental data and combine it with Inverse Reinforcement Learning (IRL) to infer the true exploitation rewards, ultimately obtaining an optimal policy that matches empirical real-data statistics. 
\subsubsection{Data preprocessing}
We utilize 30 real-world 2D trajectories of \textit{Chlamydomonas} swimming under optical fiber illumination. To translate this raw tracking data into a discrete-time POMDP format, we first reconstructed the absolute time axis from camera timestamps to compute instantaneous translational speeds $|v|$ and wrapped angular velocities $\omega$. We then applied a labeling rule: intervals were classified as a \textit{tumble} if their speed fell below a robust low-speed threshold (the 0.05 $\times$ global speeds) and their angular velocity exceeded a high-turn threshold (derived via K-Means clustering on $|\omega|$). All other intervals were labeled as \textit{runs}.

As the original camera frame intervals are irregular, we projected the trajectories onto a uniform time grid ($\Delta t$=0.125,s) via linear interpolation. The discrete action labels were mapped to this uniform grid using a majority vote based on temporal overlap. This pipeline effectively mitigates interpolation-induced drift and yields a standardized sequence of state-action-observation tuples for downstream inference.

To further evaluate the generalization of our learned models, we performed a strict train-test split on the dataset. Out of the 30 processed trajectories, we randomly reserved 4 trajectories as a held-out test set strictly for downstream benchmarking. The remaining 26 trajectories were exclusively utilized for training the POMDP policy and performing IRL.

\subsubsection{Policy learning \& benchmarking}
To recover the organism's navigational strategy, we decompose the learning process into 2 phases: (1) joint learning of the observation policy, kinematics, and POMDP emission kernels, and (2) IRL to extract the underlying exploitation reward and benchmark against objective models.

\paragraph{Phase 1: joint policy learning}
In our previous idealized simulations, we assumed tumbling agents reorient instantaneously in place ($c_{\text{tum}} = 0$). However, real cells exhibit some continuous translational displacement even while reorienting. To rigorously reflect this, we generalize our kinematics for the data-driven pipeline by introducing learnable speed scales for both actions: $c_{\text{run}}$ and $c_{\text{tum}}$ (where typically $0 < c_{\text{tum}} \le c_{\text{run}}$). 

We train a 2-layer MLP policy $\pi_\theta(a|o)$ mapping one-hot observations to action probabilities, jointly with the kinematic scales and the POMDP emission kernel $Z(o|h)$. The training minimizes a combined objective:
\begin{equation}
\mathcal{L}_{\text{Phase1}} = L_{\text{kin}} + \beta_{\text{obs}} L_{\text{obs}} + \alpha_Z \text{KL}(Z \| Z_0)
\end{equation}
where $L_{\text{obs}}$ maximizes the transition likelihood, $\text{KL}(Z \| Z_0)$ regularizes the emission kernel against the biophysical prior $Z_0$, and $L_{\text{kin}}$ is the kinematic reconstruction loss:
\begin{equation}
L_{\text{kin}} = \mathbb{E}_t \left[ \Delta t \left\{ (1 - p_{\text{tum}})\ell_{\text{run}} + p_{\text{tum}}\ell_{\text{tum}} \right\} \right]
\end{equation}
Here, $\ell_{\rm run/tum}$ represents the mean squared error between the predicted step displacement under action and the actual trajectory step. After training, the model successfully extracted biological speeds ($c_{\text{run}} \approx 18.8,px/s \ge c_{\text{tum}}\approx 0.03,px/s > 0$). Since this supervised policy primarily fits physical displacements rather than exploratory motives (producing almost zero tumbles when in the POMDP), we freeze these parameters and proceed to Phase 2.

\paragraph{Phase 2: IRL \& baseline benchmarking}
However, the purely supervised policy in phase 1 failed to capture the exploratory run-and-tumble alternation, producing almost zero tumbles. To uncover the latent functional objective driving this behavior, we apply Inverse Reinforcement Learning (IRL) that recovers the reward vector $r$ to reproduce real-data behavioral patterns with fixed learned parameters~\citep{10.5555/1953048.2021028,djeumou2021taskguidedinversereinforcementlearning}.

Crucially, to avoid the identifiability problem in reward learning \textemdash where intrinsic exploration bonuses and extrinsic task rewards can confound one another \textemdash we intentionally set the curiosity weight to zero ($\lambda=0$) during the IRL phase. This isolates the extrinsic motivation, ensuring that the recovered reward vector $r \in \mathbb{R}^n$ purely reflects the organism's physical preference for spatial alignment. We freeze the learned kernels and isolate this exploitation value function:
\begin{equation}
    Q_r(o, a) = r^\top (T_a b^+_t)
\end{equation}
With the extrinsic goal $r$ strictly anchored, the organism's inherent exploratory drive is captured macroscopically via a soft Boltzmann policy, governed by a temperature hyperparameter $\tau$:
\begin{equation}
\pi_r(a \mid o) = \frac{\exp(Q_r(o, a)/\tau)}{\sum_{a'} \exp(Q_r(o, a')/\tau)}
\end{equation}
By computing an empirical expert distribution $P_{\text{exp}}(a \mid o)$ from the real data via Laplacian smoothing, we recover the optimal reward $r^*$ by minimizing the cross-entropy loss with $L_2$ regularization:
\begin{equation}
\min_r \mathbb{E}_o \left[ -\sum_a P_{\text{exp}}(a \mid o) \log \pi_r(a \mid o) \right] + \lambda_r \|r\|_2^2
\end{equation}
The recovered reward $r^*$ successfully assigns higher values to hidden states directly facing the light, confirming our theoretical exploitation design~\citep{ziebart2008maximum}.

\subparagraph{Benchmarking: objective vs. subjective models}
To fairly evaluate our data-driven subjective POMDP, we benchmark it against the objective models. The \textit{Standard SSA} treats tumbling as an instantaneous event without displacement ($c_{\text{tum}} = 0$), serving as our classical theoretical baseline. However, since our Phase 1 learning revealed a non-zero tumbling speed ($c_{\text{tum}} > 0$), we explicitly introduce a \textit{Modified CTMC} objective baseline alongside the \textit{Standard} objective model. In the Modified CTMC version, the agent does not merely spin in place; instead, during a tumble, it experiences continuous translational displacement: $\Delta x = \Delta t \cdot c_{\text{tum}} \hat{u}_{\text{tum}}$, where $\hat{u}_{\text{tum}} \sim \text{Unif}(\mathbb{S}^1)$. 

To quantitatively compare performance, we evaluate the alignment cosine between the agent's heading $v_t$ and the true light source $\mu$:
\begin{equation}
\label{eq:align}
\cos \theta_{\text{align}} = \frac{v_t \cdot (\mu - x_t)}{\|v_t\| \|\mu - x_t\|}
\end{equation}

\begin{figure}[h]
  \centering
    \includegraphics[width=\linewidth]{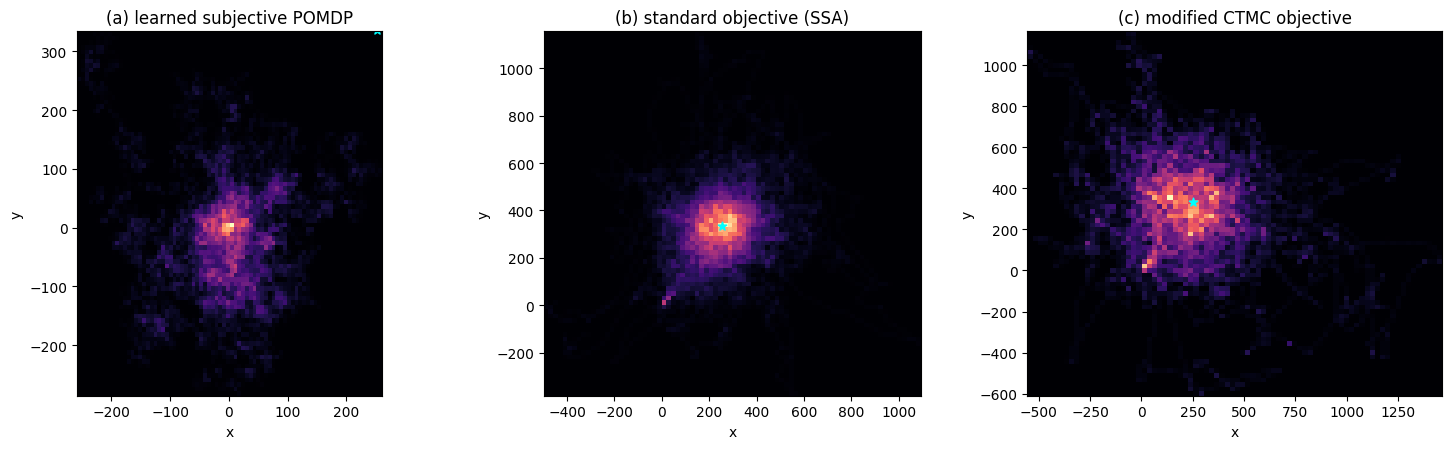}
  \caption{Spatial occupancy density (30 trajectories, fixed initial position; light source center) for
(a) learned subjective POMDP, (b) standard objective SSA, (c) modified CTMC.}
  \label{fig:traj-bench}
\end{figure}

Figure \ref{fig:traj-bench} shows the spatial occupancy density of 30 simulated trajectories (same fixed start and light-source center) for the learned subjective POMDP and the two objective baselines (idealized standard SSA \& modified CTMC), where brighter regions indicate more frequently visited locations. All three concentrate occupancy around the light source, confirming successful phototaxis, but the learned POMDP spreads its occupancy more broadly than the two objective baselines. Because such occupancy maps are only qualitative, we quantify the behavior below using alignment-distribution metrics rather than the spatial plots alone.

Beyond spatial occupancy, we compare the full distribution of the alignment cosine $\cos\theta_{\mathrm{align}}$ (Eq.~\ref{eq:align}) across models. To quantify this congruence, we computed the Wasserstein-1 distance ($W_1$) and the Kolmogorov–Smirnov statistic ($D$), summarized in Table~\ref{tab:metrics}. The learned subjective policy is closest to the Modified CTMC ($W_1\approx 0.0187$, $D\approx 0.0156$) and remains close to the held-out real data ($W_1\approx 0.0458$, $D\approx 0.0536$), consistent with a memoryless POMDP recovering gradient-ascent behavior under realistic fluid kinematics ($c_{\mathrm{tum}}>0$). Since the held-out set contains only four trajectories ($n=2773$ alignment samples), these distances are descriptive diagnostics rather than formal generalization guarantees. 95\% bootstrap intervals over alignment samples are: Real vs.\ Policy $W_1\!\in\![0.026,0.072]$, $D\!\in\![0.038,0.074]$; Policy vs.\ Modified CTMC $W_1\!\in\![0.014,0.024]$,
$D\!\in\![0.013,0.020]$.

\begin{table}[htbp]
\centering
\resizebox{0.9\columnwidth}{!}{%
\begin{tabular}{lcc}
\toprule
\textbf{Comparison Pair} & \textbf{Wasserstein-1 ($W_1$)} & \textbf{KS Statistic ($D$)} \\ \midrule
Policy vs. Modified CTMC & $\mathbf{0.0187}$ & $\mathbf{0.0156}$ \\
Policy vs. Standard SSA  & 0.0441 & 0.0387 \\
Policy vs. real data     & 0.0458 & 0.0536 \\
\bottomrule
\end{tabular}%
}
\caption{\textbf{Statistical distances of alignment distributions.} The metrics evaluate the congruence between the empirical real data (held-out test set), the objective baselines, and the learned subjective POMDP policy. Lower Wasserstein-1 ($W_1$) and Kolmogorov-Smirnov ($D$) values indicate higher distributional similarity.}
\label{tab:metrics}
\end{table}


Because this real-data column is computed exclusively on the four held-out trajectories---never used for Phase-1 training or IRL---it is a genuine held-out comparison rather than an in-sample goodness-of-fit, and therefore supports, rather than proves, that the IRL-recovered reward captures biologically relevant alignment preferences. Despite this close spatial alignment, the action frequencies diverge markedly, motivating the exploration--exploitation analysis below.


\paragraph{Exploration-exploitation trade-off}

Despite the excellent alignment distribution, a crucial discrepancy emerged in action frequencies. As shown in Table \ref{tab:policy-tumble}, integrating realistic kinematics ($c_{\text{tum}}>0$) into the Modified CTMC inherently raises the tumble ratio to $0.130$ compared to the idealized Standard SSA ($0.038$). Because reorienting while moving dilutes turning efficiency, the agent must tumble more frequently to achieve the same alignment. 

Our data-driven POMDP captures this kinematic reality, which explains why the initial IRL soft policy ($\tau=1.0$) overestimates the tumbling frequency ($0.503$). To bridge this gap toward the real experimental data ($0.027$), we calibrated the temperature hyperparameter to $\tau^* \approx 0.007$. The temperature was calibrated post hoc by minimizing the absolute mismatch between the
simulated and empirical tumble ratios,
\[
\tau^\ast=\arg\min_\tau\bigl|\rho_{\mathrm{sim}}(\tau)-\rho_{\mathrm{data}}\bigr|,
\]
evaluated with $8$ runs of $800$ steps per $\tau$ (target $\rho_{\mathrm{data}}=p_{\mathrm{real}}$).
This changes only the stochasticity of action \emph{sampling}: the IRL-recovered reward $r^\ast$
(learned at $\lambda=0$) explains the alignment \emph{preferences}, whereas $\tau$ is a behavioral
sharpness parameter that matches action \emph{frequency}, not a re-fit of the reward. This sharpens the policy, successfully reducing the simulated tumble ratio to $0.0121$, bringing it closer to the empirical value while still leaving a measurable gap.

\begin{table}[ht]
  \centering

  \resizebox{\columnwidth}{!}{%
  \begin{tabular}{lll}
    \hline
    Model / Data                          & Setting                 & Tumble ratio \\
    \hline
    real data                             & ---                     & $0.027$      \\
    objective(standard)                        & instantaneous reorient. & $0.038$      \\
    objective(modified)                        & run-while-tumble CTMC   & $0.13$       \\
    Learned subjective (before IRL)                  & -            & $0$      \\
    Learned subjective (after IRL)                  & $\tau = 1.0$            & $0.503$      \\
    Learned subjective (calibrated)     & $\tau^\star=0.007$   & $0.0121$     \\
    \hline
  \end{tabular}%
  }
\caption{Tumble ratios for real data, objective baselines and learned subjective models.}
  \label{tab:policy-tumble}
\end{table}

\begin{figure}[htbp]
  \centering
  \begin{subfigure}{0.48\linewidth}
    \centering
    \includegraphics[width=\linewidth]{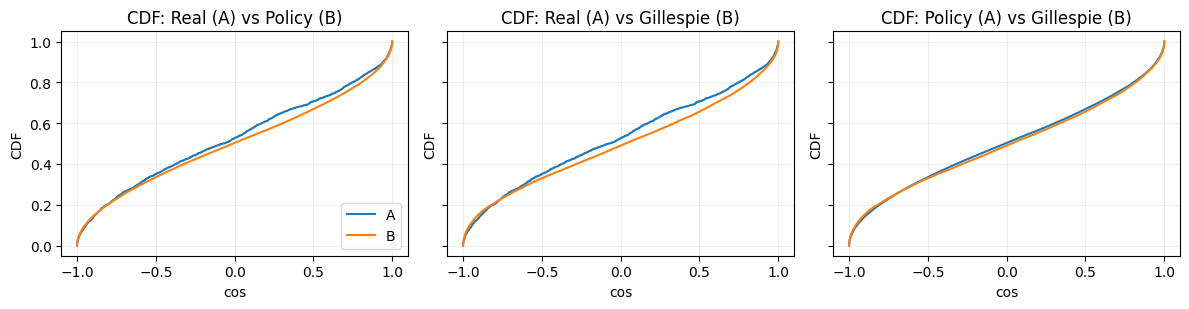}
    \caption{CDFs: data, SSA, learned policy ($\tau$).}
  \end{subfigure}\hfill
  \begin{subfigure}{0.48\linewidth}
    \centering
    \includegraphics[width=\linewidth]{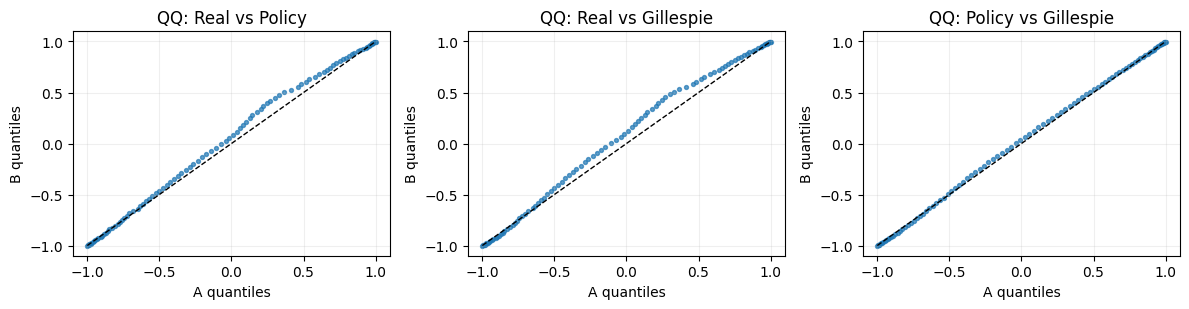}
    \caption{Q--Q plots: data, SSA, learned policy ($\tau$).}
    
  \end{subfigure}
   \begin{subfigure}{0.48\linewidth}
    \centering
    \includegraphics[width=\linewidth]{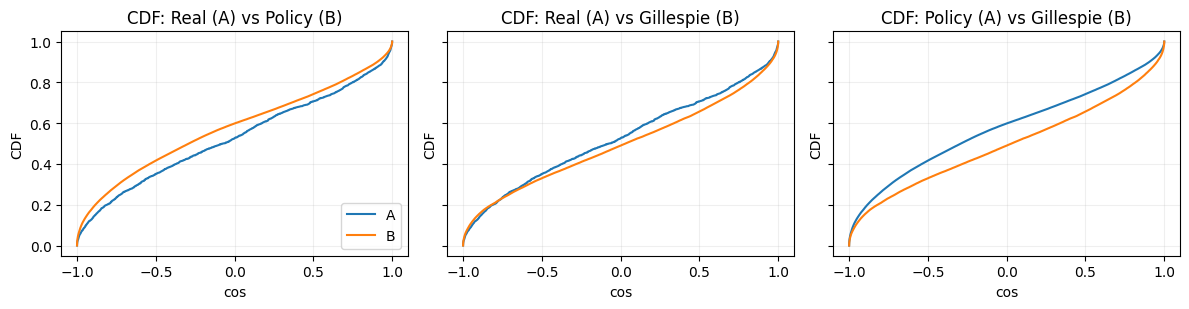}
    \caption{CDFs: data, SSA, learned policy ($\tau^\star$).}
  \end{subfigure}\hfill
  \begin{subfigure}{0.48\linewidth}
    \centering
    \includegraphics[width=\linewidth]{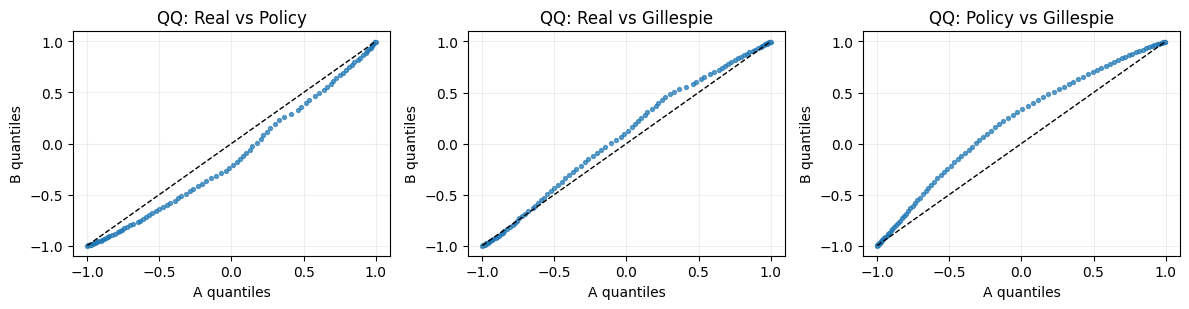}
    \caption{Q--Q plots: data, SSA, learned policy ($\tau^\star$).}
  \end{subfigure}
  \caption{Empirical CDFs and Q--Q plots of the alignment cosine
  $\cos\theta_{\text{align}}$ (heading vs.\ light direction) for real data, SSA baselines
  and learned POMDP policies, with the original temperature $\tau$ (top) and the calibrated
  temperature $\tau^\star$ (bottom).}
  \label{fig:alignment-summary}
\end{figure}

We compare the resulting alignment-cosine distributions using empirical cumulative distribution functions (CDFs) and quantile--quantile (Q--Q) plots against the real data and SSA baselines (Fig.~\ref{fig:alignment-summary}). Calibrating the temperature $\tau$ suppresses excess tumbling but slightly broadens the alignment distribution (Fig.~\ref{fig:alignment-summary}c, d).

However, suppressing the tumble frequency naturally comes at a cost: without frequent reorientations to correct drifting errors, the alignment distribution broadens. Quantitatively, the alignment error against the real data measurably increased (with $W_1$ rising from $0.0458$ to $\mathbf{0.1006}$, and the KS statistic $D$ increasing from $0.0536$ to $\mathbf{0.0797}$).

We interpret this shift cautiously. Lowering $\tau$ reduces the tumble frequency toward the
empirical value but broadens the alignment distribution, so a single scalar temperature
cannot simultaneously match action frequency and alignment accuracy. Rather than a fitted
``biological trade-off,'' this reflects a tension between locomotive efficiency and
reorientation accuracy under run-while-tumble kinematics ($c_{\mathrm{tum}}>0$); a richer
policy (e.g.\ uncertainty-dependent tumbling) may be required to close the residual gap.

\section{Discussion}

By bridging abstract algorithmic reinforcement learning with physically realizable mass-action kinetics, our data-driven POMDP framework demonstrates that macroscopic phototactic behavior can be replicated by subjective, memoryless inference coupled to mass-action-compatible dynamics. Our results suggest a reinterpretation of run-and-tumble behavior as an information-seeking strategy that helps resolve perceptual ambiguity during navigation. In this view, tumbling is not merely stochastic noise but a functional, curiosity-driven mechanism for actively sampling the environment when directional cues are uncertain. By linking this behavioral dynamics to biochemical reaction networks, we show how such exploratory processes can be physically realized through intracellular molecular interactions. This provides a bridge between abstract descriptions of adaptive behavior and the underlying biochemical mechanisms that can implement them. 

This work situates itself among recent efforts exploring microbial navigation through advanced computational paradigms. Related work has explored run-and-tumble navigation in terms of reinforcement learning~\citep{pramanik2025run} and active inference frameworks emphasizing curiosity-driven exploration \cite{tschantz2020learning}, particularly in the context of chemotaxis. Other studies have shown that the structure of biochemical pathways involved in chemotaxis or phototaxis may be inferred from the tumbling dynamics \cite{lei2025identification}. Our approach complements these perspectives by explicitly demonstrating how information-seeking dynamics can be implemented through modular biochemical reaction networks compatible with mass-action kinetics.

From an artificial life perspective, these results highlight how adaptive exploratory behavior can emerge from relatively simple biochemical circuits interacting with uncertain environments. Such mechanisms illustrate how minimal biological systems may realize  sophisticated information-processing strategies without requiring complex neural architectures. Behavioral variability, historically dismissed as noise, may therefore play a constructive role in shaping effective exploration.

Looking forward, this framework provides a foundation for several possible avenues. The tumble ratios reported here are trajectory-level aggregates. An exploratory analysis of tumble fraction versus distance to the light source indicates that this global statistic can conceal a spatial dependence; conditioning tumble probability on distance, alignment error, and posterior uncertainty is a natural refinement left to future work. Beyond the Shannon mutual information used here, deformed (Tsallis) versions of the Kullback--Leibler divergence are rational functions~\citep{PhysRevE.58.1442,e13091694} and can therefore be realized as steady states of a CRN; implementing such a $q$-deformed information gain is a natural extension. This framework should also be extended to richer environments, such as multiple or dynamically varying light sources, where more complex collective exploratory strategies may emerge. Further investigation of the biochemical modules may also support the design of synthetic control mechanisms for phototactic microorganisms. Finally, these principles could inspire embodied implementations in minimal robotic systems, where simple sensorimotor loops reproduce biologically inspired exploration and active sensing strategies in the physical world.


\section{Acknowledgements}
We thank Pierre Bessière and Jacques Droulez for advice on the observation design and evaluation protocol, and for discussions that improved our evaluation setup and presentation. We also thank Sorbonne University and Sony CSL for supporting this research.

This internship was funded by AMIES – PEPS project "\textit{Generative models of chemotaxis and phototaxis}".

The authors used generative AI tools for language editing and organizational assistance during manuscript preparation. All scientific content, analyses, and conclusions were reviewed and verified by the authors.

\footnotesize
\bibliographystyle{apalike}
\bibliography{references} 

\end{document}